# Circle detection on images using Learning Automata

Erik Cuevas, Fernando Wario, Daniel Zaldivar and Marco Pérez-Cisneros

Departamento de Ciencias Computacionales

Universidad de Guadalajara, CUCEI

Av. Revolución 1500, Guadalajara, Jal, México

{erik.cuevas, fernando.wario, daniel.zaldivar, marco.perez}@cucei.udg.mx

## Abstract

Circle detection over digital images has received considerable attention from the computer vision community over the last few years devoting a tremendous amount of research seeking for an optimal detector. This article presents an algorithm for the automatic detection of circular shapes from complicated and noisy images with no consideration of conventional Hough transform principles. The proposed algorithm is based on Learning Automata (LA) which is a probabilistic optimization method that explores an unknown random environment by progressively improving the performance via a reinforcement signal (objective function). The approach uses the encoding of three non-collinear points as a candidate circle over the edge image. A reinforcement signal (matching function) indicates if such candidate circles are actually present in the edge map. Guided by the values of such reinforcement signal, the probability set of the encoded candidate circles is modified through the LA algorithm so that they can fit to the actual circles on the edge map. Experimental results over several complex synthetic and natural images have validated the efficiency of the proposed technique regarding accuracy, speed and robustness.

## 1. Introduction

The problem of detecting circular features is very important for image analysis, in particular for industrial applications such as automatic inspection of manufactured products and components, aided vectorization of drawings, target detection, etc. [1]. Circular Hough transform [2] is arguably the most common technique for circle detection in digital images. A typical Hough-based approach employs an edge detector to infer locations and radius values. Averaging, filtering and histogramming of the transformed space are subsequently applied. The approach demands a large storage space as 3-D cells to store operational parameters ($x$, $y$, $r$), seriously constraining the overall performance to low processing speeds. In Hough Transform methods, circle's parameters are poorly defined under noisy conditions [3] yielding a longer processing time which constraints their application. In order to overcome such problems, researchers have proposed new Hough transform-based (HT) approaches such as the probabilistic HT [4], the randomized HT (RHT) [5] and the fuzzy HT (FHT) [6]. In [7], Lu & Tan proposed a novel approach based on RHT called Iterative Randomized HT (IRHT) that achieves better results on complex images and noisy environments. Such implementation applies iteratively the RHT to a given region of interest which has been previously defined from the latest estimation of ellipse/circle parameters.

Alternatively to the Hough Transform, the shape recognition problem in computer vision has also been handled with optimization methods. In particular, Genetic Algorithms (GA) have recently been used for important shape detection tasks. For instance, Roth and Levine have proposed the use of GA for extracting geometrical primitives [8]. Lutton *et al* have developed an improvement of the aforementioned method in [9] while Yao *et al* have proposed a multi-population GA to detect ellipses [10]. In [11], GA have been used for template matching despite the available pattern has been modified by an unknown affine transformation. Ayala–Ramirez *et al* have presented a GA based circle detector in [12] which is able to detect multiple circles on real images but failing frequently on imperfect circles.

This paper assumes the circle detection problem as an optimization algorithm and develops an alternative approach based on Learning Automata (LA) [13-15]. LA is an adaptive decision making method that operate at an unknown random environment while progressively improving their performance via a learning process. A probability density function is defined over the parameter space where each parameter (or parameters in case of a multidimensional problem) represents an action which is applied to a random environment. The corresponding response from the environment, which is also known as reinforcement signal, is used by the automata to update the probability density function at each stage in order to select its next action. The procedure continues until an optimal action is defined.

The main motivation behind the use of LA refers to its abilities as global optimizer for multimodal surfaces. Optimization techniques based on Learning Automata (LA) fall into the random search class.

The distinguishing characteristic of automata-based learning is that the searching for the optimal parameter vector is performed within the space of probability distributions which has been defined over the parameter space rather than over the parameter space itself [16]. Therefore LA has been employed to solve different sorts of engineering problems, for instance, pattern recognition [17], adaptive control [18], signal processing [19], power systems [20] and computer vision [21]. Other interesting applications for multimodal complex function optimization based on the LA have been proposed in [19, 22, 23, 24], yet showing that their performance is comparable to (GA) [23].

This paper presents an algorithm for the automatic detection of circular shapes from complicated and noisy images with no consideration of conventional Hough transform principles. The proposed algorithm LA requires the probability of three encoded non-collinear edge points as candidate circles (actions). A reinforcement signal indicates if such candidate circles are actually present in the edge-only image. Guided by the values of such performance evaluation function, the probability set of the encoded candidate circles is modified using the LA algorithm so that they can fit into the actual circles (optimal action) in the edge map. The approach generates a sub-pixel circle detector which can effectively identify circles in real images despite circular objects exhibiting a significant occluded portion. Experimental evidence shows its effectiveness for detecting circles under different conditions. A comparison to other state-of-the-art methods such as the GA algorithm [12] and the Iterative Randomized Hough Transform approach (IRHT) [7] on multiple images has demonstrated the improved performance of the proposed method.

The paper is organized as follows: Section 2 provides a brief outline of LA theory while Section 3 presents the LA-based circle detector. Section 4 shows the results of applying the LA algorithm for circle recognition under several image conditions and section 5 presents a performance comparison between the proposed method and other relevant techniques reported in the literature. Finally Section 6 discusses on some relevant conclusions.

**2. Learning automata (LA)**

LA is a finite state machine that interacts to a stochastic environment while learning about an optimal action which is conditioned by the environment through a learning process. Figure 1a shows the typical

LA system architecture. At each instant *k*, the automaton **B** selects probabilistically an action $B_{current}$ from the set of *n* actions. After applying such action to the environment, a reinforcement signal $\beta(B_{current})$ is provided through the evaluation function. The internal probability distribution $\mathbf{p}(k) = \{p_1(k), p_2(k), \ldots, p_n(k)\}$ is updated whereby actions that achieve desirable performance are reinforced via an increased probability while those not-performing actions are penalized or left unchanged depending on the particular learning rule which has been employed. The procedure is repeated until the optimal action $B_{optimal}$ is found. From an optimization-like perspective, the action with the highest probability (optimal action) corresponds to the global minimum as it is demonstrated by rigorous proofs of convergence available in [16] and [25].

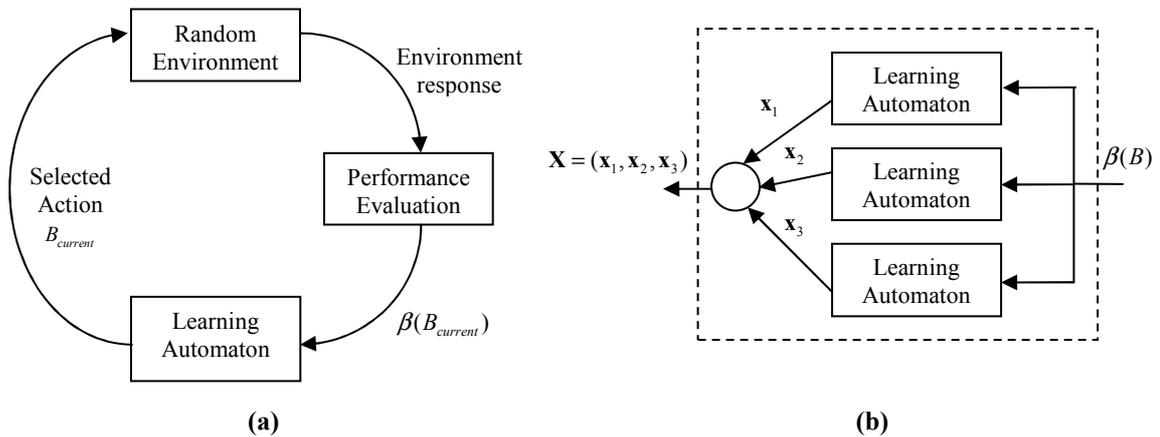

**Figure 1.** (a) The reinforcement learning system and (b) its parallel connected automata.

The operation of a LA during one iteration consists of two basic functions:

(a) Probability updating. Based on the environmental response to the selected action $\beta(B_{current})$, the automaton modifies the probability distribution $\mathbf{p}(k)$ over the set of actions to $\mathbf{p}(k+1)$.

(b) Action selection. Based on the new probability distribution $\mathbf{p}(k+1)$, the automaton selects a new action $B_{new}$ that is applied to the environment.

A wide variety of updating rules for probabilities have been reported in the literature for LA. One of the most widely used algorithms is the linear reward/inaction ($L_{RI}$) scheme, which has shown effective convergence properties (see [22]). Considering an automaton **B** with $n$ different actions, $B_r$ represents the action $r$ of a set of $n$ possible actions. As a response to an action $B_r$, at time step $k$, the probabilities $\mathbf{p}(k)$ are updated as follows:

$$p_r(k+1) = p_r(k) + \theta \cdot \beta(B_r) \cdot (1 - p_r(k)) \tag{1}$$

$$p_q(k+1) = p_q(k) - \theta \cdot \beta(B_r) \cdot p_q(k), \text{ if } q \neq r$$

with $\theta$ being the learning rate and $0 < \theta < 1$, $\beta(\cdot) \in [0,1]$ the reinforcement signal whose value $\beta(\cdot) = 1$ indicates the maximum reward and $\beta(\cdot) = 0$ signals a null reward considering $r, q \in \{1, \ldots, n\}$. Using the $L_{RI}$ scheme, the probability of successful actions will increase until they become close to unity.

On the other hand, a uniformly distributed pseudo-random number $z$ is generated in the range [0, 1] for selecting the new action $B_{new} \in (B_1, B_2, \ldots B_n)$ after considering the probability density function $\mathbf{p}(k+1)$. Thus action $l$ is chosen following:

$$\sum_{h=1}^{l} p_h(k+1) > z \tag{2}$$

Therefore, the chosen action $B_l$ triggers the environment which responds through feedback $\beta(B_l)$ and continues the loop. As stop criteria, the LA algorithm is constraint to a cycle number that is usually half of the number of actions considered by the automaton. Once the cycle number has been reached, the action holding the best probability value is taken as the solution $B_{optimal}$.

In order to solve multidimensional problems, the learning automata can also become connected to a parallel setup (see Figure 1b). Each automaton operates with a simple parameter while its concatenation allows working within a multidimensional space. There is no inter-automata communication as the only joining path is through the environment. In [13], it is shown how discrete stochastic learning automata can be used to determine the global optimum for problems with multi-modal surfaces.

### 3. Circle detection using LA

*3.1. Data preprocessing*

In order to apply the LA circle detector, candidate images must be preprocessed first by the well-known Canny algorithm which yields a single-pixel edge-only image. Then, the $(x_i, y_i)$ coordinates for each edge pixel $p_i$ are stored inside the edge vector $P_t = \{p_1, p_2, \ldots, p_{N_t}\}$, with $N_t$ being the total number of edge pixels. Following the RHT technique in [12], only a representative percentage of edge points (around 5%) are considered for building the new vector array $P = \{p_1, p_2, \ldots, p_{N_p}\}$, where $N_p$ is the number of edge pixels randomly selected from $P_t$.

*3.2. Action representation*

In order to construct each action $C_i$ (circle candidate), the indexes $i_1$, $i_2$ and $i_3$, which represent three edge points previously stored inside the vector $P$, must be grouped assuming that the circle's contour connects them. Therefore, the circle $C_i = \{p_{i_1}, p_{i_2}, p_{i_3}\}$ passing over such points may be considered as a potential solution for the detection problem. Considering the configuration of the edge points shown by Figure 2, the circle center $(x_0, y_0)$ and the radius $r$ of $C_i$ can be characterized as follows:

$$(x - x_0)^2 + (y - y_0)^2 = r^2 \qquad (3)$$

where $x_0$ and $y_0$ are computed through the following equations:

$$x_0 = \frac{\det(\mathbf{A})}{4((x_{i_2} - x_{i_1})(y_{i_3} - y_{i_1}) - (x_{i_3} - x_{i_1})(y_{i_2} - y_{i_1}))}, \; y_0 = \frac{\det(\mathbf{B})}{4((x_{i_2} - x_{i_1})(y_{i_3} - y_{i_1}) - (x_{i_3} - x_{i_1})(y_{i_2} - y_{i_1}))} \qquad (4)$$

with $\det(\mathbf{A})$ and $\det(\mathbf{B})$ representing determinants of matrices $\mathbf{A}$ and $\mathbf{B}$ respectively, considering:

$$A = \begin{bmatrix} x_{i_2}^2 + y_{i_2}^2 - (x_{i_1}^2 + y_{i_1}^2) & 2 \cdot (y_{i_2} - y_{i_1}) \\ x_{i_3}^2 + y_{i_3}^2 - (x_{i_1}^2 + y_{i_1}^2) & 2 \cdot (y_{i_3} - y_{i_1}) \end{bmatrix} \quad B = \begin{bmatrix} 2 \cdot (x_{i_2} - x_{i_1}) & x_{i_2}^2 + y_{i_2}^2 - (x_{i_1}^2 + y_{i_1}^2) \\ 2 \cdot (x_{i_3} - x_{i_1}) & x_{i_3}^2 + y_{i_3}^2 - (x_{i_1}^2 + y_{i_1}^2) \end{bmatrix},\quad(5)$$

the radius $r$ can therefore be calculated using:

$$r = \sqrt{(x_0 - x_d)^2 + (y_0 - y_d)^2},\quad(6)$$

where $d \in \{i_1, i_2, i_3\}$, and $(x_d, y_d)$ are the coordinates of any of the three selected points which define the action $C_i$. Figure 2 illustrates main parameters defined by Equations (3)-(6).

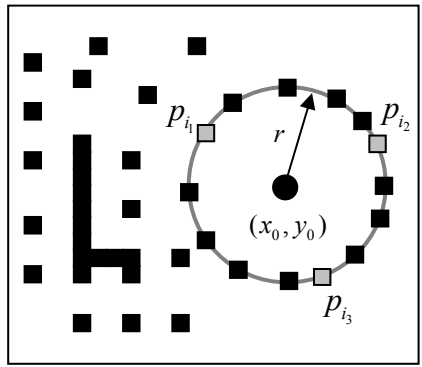

**Fig. 2.** Circle candidate (action) formed from the combination of points $p_{i_1}$, $p_{i_2}$ and $p_{i_3}$.

The shaping parameters for the circle, $[x_0, y_0, r]$ can be represented as a transformation $T$ of the edge vector indexes $i_1$, $i_2$ and $i_3$.

$$[x_0, y_0, r] = T(i_1, i_2, i_3) \quad(7)$$

The total number of actions $n_{all}$ is generated considering all feasible combinations of $P$. After calculating the circle parameters $(x_0, y_0, r)$ using Eq. (7), only the actions whose radii fall into a determined range are considered. The allowed range is defined as $8 < r < \max(I(\text{columns})/2, I(\text{rows})/2)$ where $I(\text{columns})$ and $I(\text{rows})$ represent the maximum number of columns and rows respectively inside the image. Moreover, actions that correspond to circles already marked are eliminated. Hence, the final number of actions $n_c$, represents the resulting solution set.

The LA solution is based on tracking the probability evolution for each circle candidate, also known as actions, as they are modified according to their actual affinity. Such affinity is computed using an objective function which determines if a circle candidate is actually present inside the image. Following a number of cycles, the circle candidate showing the highest probability value is assumed as a circle actually present in the image.

Although the HT based methods for circle detection also use three edge points to cast one vote for a potential circular shape in the parameter space, they require huge amounts of memory and longer computational times to reach a sub-pixel resolution. On the contrary, the LA method employs an objective function yielding improvement at each generation step, discriminating among non-plausible circles and avoiding unnecessary testing of certain image points. However, both methods require a compulsory evidence-collecting step for future iterations.

*3.3 Performance evaluation function $\beta(\ )$*

In order to model the environment's reaction after an action $C_i$ is applied, the circumference coordinates of the circle candidate $C_i$ are calculated as a virtual shape which must be validated, i.e. if it really exists in the edge image. The circumference coordinates are grouped within the test set $S_i = \{s_1, s_2, \ldots, s_{N_s}\}$, with $N_s$ representing the number of points over which the existence of an edge point, corresponding to $C_i$, should be verified.

The test $S_i$ is generated by the midpoint circle algorithm (MCA) [26] which is a well-known algorithm to determine the required points for drawing a circle. MCA requires as inputs only the radius $r$ and the center point $(x_0, y_0)$ considering only the first octant over the circle equation $x^2 + y^2 = r^2$. It draws a curve starting at point $(r, 0)$ and proceeds upwards-left by using integer additions and subtractions. The MCA aims to calculate the required points $S_i$ in order to represent a circle candidate. Although the algorithm is considered as the quickest providing a sub-pixel precision, it is important to assure that points lying outside the image plane must not be considered in $S_i$.

The reinforcement signal $\beta(C_i)$ represents the matching error produced between the pixels $S_i$ of the circle candidate $C_i$ (action) and the pixels that actually exist in the edge-only image, yielding:

$$\beta(C_i) = \frac{\sum_{h=1}^{N_s} E(s_h)}{N_s} \quad (8)$$

where $E(s_h)$ is a function that verifies the pixel existence in $s_h$, being $s_h \in S_i$ and $N_s$ is the number of pixels lying over the perimeter and corresponding to $C_i$, currently under testing. Hence the function $E(s_h)$ is defined as:

$$E(s_h) = \begin{cases} 1 & \text{if the test pixel } s_h \text{ is an edge point} \\ 0 & \text{otherwise} \end{cases} \quad (9)$$

A value of $\beta(C_i)$ near to unity implies a better response from the "circularity" operator. Figure 3 shows the procedure to evaluate a candidate action $C_i$ with its representation as a virtual shape $S_i$. Figure 3(a) shows the original edge map, while Figure 3(b) presents the virtual shape $S_i$ representing the action $C_i = \{p_{i_1}, p_{i_2}, p_{i_3}\}$. In Figure 3(c), the virtual shape $S_i$ is compared to the original image, point by point, in order to find coincidences between virtual and edge points. The action has been built from points $p_i$, $p_j$ and $p_k$ which are shown by Fig. 3(a). The virtual shape $S_i$, obtained by MCA, gathers 56 points ($N_s = 56$) with only 18 of them existing in both images (shown as blue points plus red points in Fig. 3(c)) and yielding: $\sum_{h=1}^{Ns} E(s_h) = 18$, therefore $\beta(C_i) \approx 0.33$.

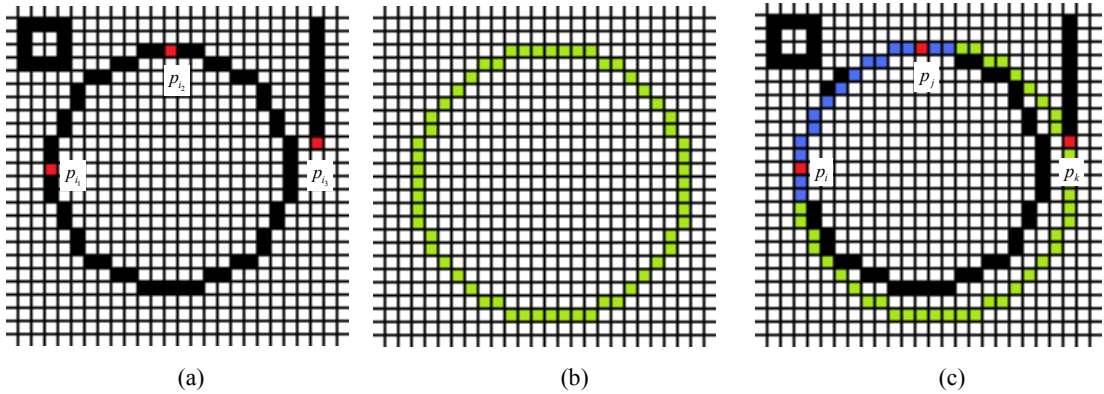

**Fig. 3.** Environment reaction to an action $C_i$: The image shown by (a) presents the original edge image while (b) portraits the virtual shape $S_i$ corresponding to $C_i$. The image in (c) shows coincidences between both images through blue or red pixels while the virtual shape is also depicted in green.

The LA algorithm is set to a pre-selected cycle limit that is usually chosen to half the number of actions ($n_c/2$) that form the automaton. There are two cases to obtain a solution (optimal action), either if one action (circle candidate) generates a matching error $\beta(\ )$ under the pre-established limit or takes the highest probability action at the end of the learning process.

*3.4. LA implementation*

Considering the image has been pre-processed by a canny filter, the LA-detector procedure can be summarized as follows:

**Step 1:** Select 5% of edge pixels to build the *P* vector and generate $n_{all}$, considering all feasible combinations.

**Step 2:** Generate $n_c$ by calculating $[x_0, y_0, r] = T(i_1, i_2, i_3)$ from $n_{all}$ selecting actions which either fall into the scope or are not repeated.

**Step 3:** Set iteration *k*=0.

**Step 4:** Initialize $\mathbf{p}(k) = \{p_1(k), p_2(k), \ldots, p_{n_c}(k)\}$ as a uniform distribution.

**Step 5:** Repeat while $k < (n_c/2)$

**Step 6:** Select a new action $C_v \in (C_1, C_2, \ldots C_{n_c})$ according to $\mathbf{p}(k)$.

  

**Step 7:** Update $\mathbf{p}(k+1)$ according to $\beta(C_v)$ using Eq.(1).

**Step 8:** Increase $k$ and jump to step 6.

**Step 9:** end of while.

**Step 10:** After $k > (n_c / 2)$, the solution $C_{optimal}$ (circle) is the highest element of $\mathbf{p}$.

## 4. Experimental results

In order to evaluate the performance of the proposed LA circle detector, several experimental tests are presented as follows:

(1) Circle localization

(2) Shape discrimination

(3) Multiple circle localization

(4) Circular approximation

(5) Occluded circles and arc detection

(6) Complex cases

Table 1 presents the experimental parameter set for the LA implementation which has been experimentally determined and kept for all test images through all experiments.

| kmax | $\theta$ |
|---|---|
| $n_c / 2$ | 0.003 |

**Table 1.** LA circle detector parameters

All the experiments are performed on a Pentium IV 2.5 GHz computer under C language programming with all images being preprocessed by the standard Canny edge-detector from the image-processing toolbox for MATLAB R2008a.

*4.1 Circle localization*

4.1.1. Synthetic images

The experimental setup includes the use of twenty synthetic images of 200x200 pixels. Each image has been generated drawing only one imperfect circle (ellipse shape), randomly located. Some images have been contaminated by adding noise to increase the complexity in the detection process. The experiment aims to detect the center of the circle position ($x$, $y$) and its radius ($r$), allowing only 50 epochs for each test image. For all the cases, the algorithm is able to detect best circle parameters despite the noise influence. The detection is robust to translation and scale keeping a reasonably low elapsed time (typically under 0.1 s). Figure 4 shows the results of the circle detection acting over a synthetic image. Figure 4b shows the detected circle as an overlay while Figure 4c presents the action's probability distribution $\mathbf{p}(k)$ with the highest probability action being represented by the highest peak.

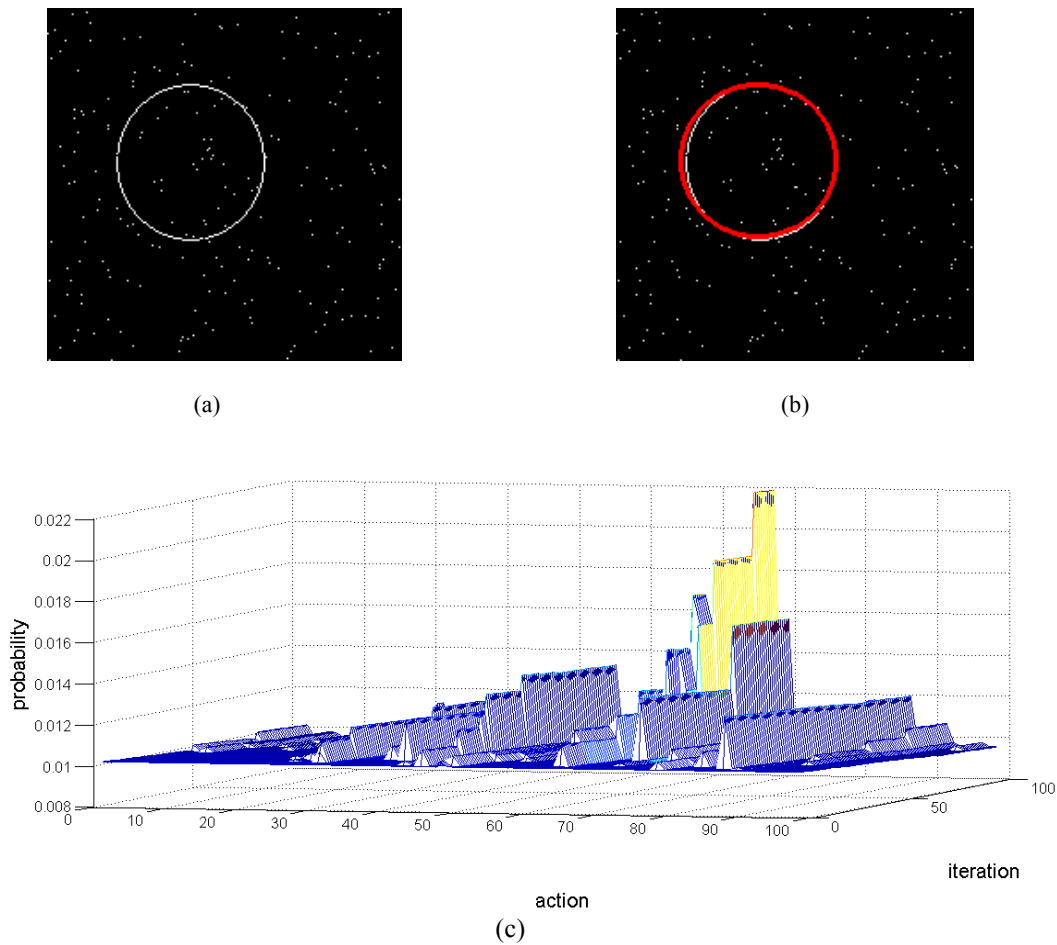

(a)          (b)

(c)

**Fig. 4.** Circle detection and the evolution of the probability density parameters. (a) Original image. (b) The detected circle is shown as an overlay, (c) parameter evolution yielding the probability density.

4.1.2. Natural images

The experiment tests the LA circle detector's performance upon real-life images. Twenty five images of 640x480 pixels are used on the test. All images have been captured by using digital camera under 8-bit color format. Each natural scene includes a circle shape among other objects. All images are preprocessed using an edge detection algorithm and then fed into the LA-based detector. Figure 5 shows a particular case from the 25 test images. Real-life images rarely contain perfect circles so the detection algorithm approximates the circle that better adapts to the imperfect circle within a noisy image. Such circle corresponds to the smallest error from the objective function $\beta(\ )$. Detection results have been statistically analyzed for comparison purposes. For instance, the detection algorithm is executed 100 times on the same image (Figure 5), yielding same parameters $x_0 = 231$, $y_0 = 301$, and $r = 149$. This indicates that the proposed LA algorithm is able to converge to a minimum solution from the objective function $\beta(\ )$.

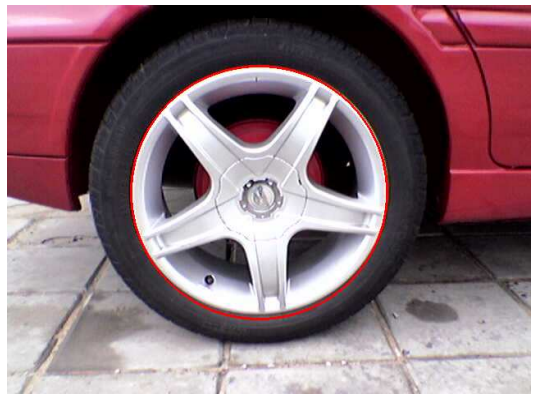

**Fig. 5.** Circle detection on a natural image. Detected circle is shown as an overlay.

*4.2. Circle discrimination tests*

4.2.1. Synthetic images

This section discusses on the algorithm's ability to detect circles despite the image featuring any other shape. Five synthetic images of 540x300 pixels are considered for the experiment. Noise has been added to all images as to increase the complexity in the detection process. Figure 6a shows a synthetic image containing different shapes including an overlapped circle, while Figure 6b presents the detected circle marked by a red overlay.

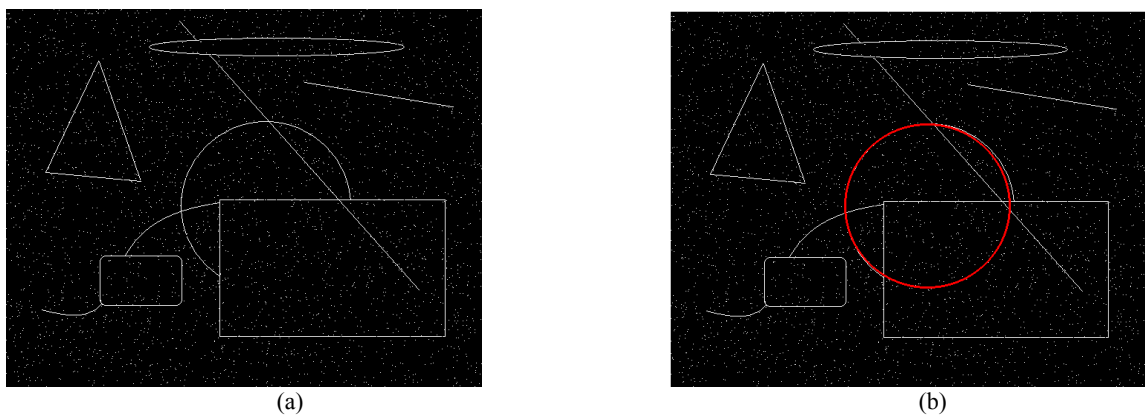

(a)          (b)

**Fig. 6.** One sample synthetic image with a variety of shapes. (a) sample input (b) the detected circle.

4.2.2. Natural images

The experiment is repeated considering real-life images. Figure 7 shows an example containing one circular shape among others.

*4.3. Multiple circle detection*

The LA circle detector is also capable of detecting several circles embedded into images. The approach is applied over the edge-only image until the first circle is detected, i.e. the $C_{optimal}$ circle holding the maximum probability value is located. That shape is thus masked (i.e. eliminated) on the primary edge-only image. Then, the LA circle detector operates again over the modified image. The procedure is repeated until the $\beta(\ )$ value reaches a minimum predefined threshold $M_{th}$ (typically 0.1). Finally, a validation over all detected circles is performed by analyzing continuity of the detected circumference segments as proposed in [27]. If none of the detected shapes satisfies the $M_{th}$ criterion, the system simply reply a negative response such as: "no circle detected".

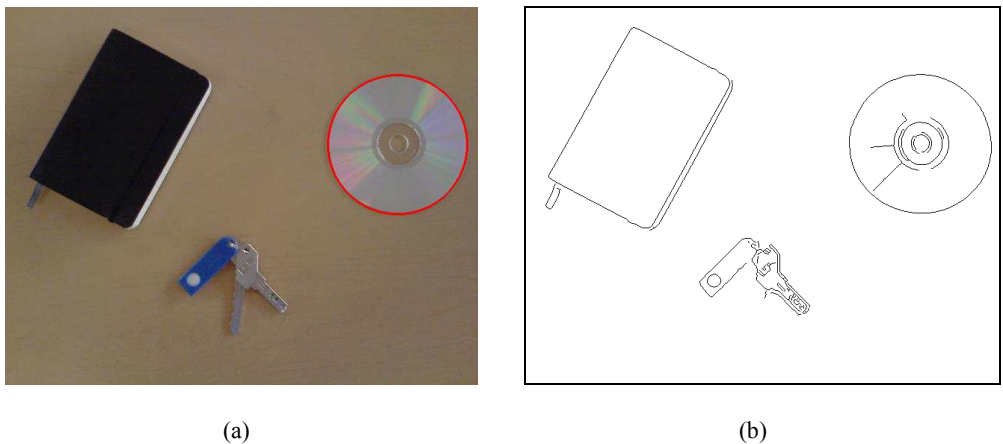

(a)          (b)

**Fig. 7.** Natural image with a variety of shapes: (a) the original image with an overlay for the detected circle and (b) the corresponding edge map.

Figure 8a shows a natural image containing several circles. For this case, the algorithm searches for the best circular shapes (greater than $M_{th}$). Figure 8b depicts the edge image after applying the Canny algorithm and prior to be fed into the LA algorithm.

*4.4 Circular approximation*

4.4.1. Synthetic images

In this paper, the circle detection process is considered to be similar to an optimization problem. Hence it is feasible to approximate other circular-like shapes by means of concatenating circles. The LA method is thus considered to detect circular patterns showing the highest probability that can be subsequently reshaped into a more complex geometry.

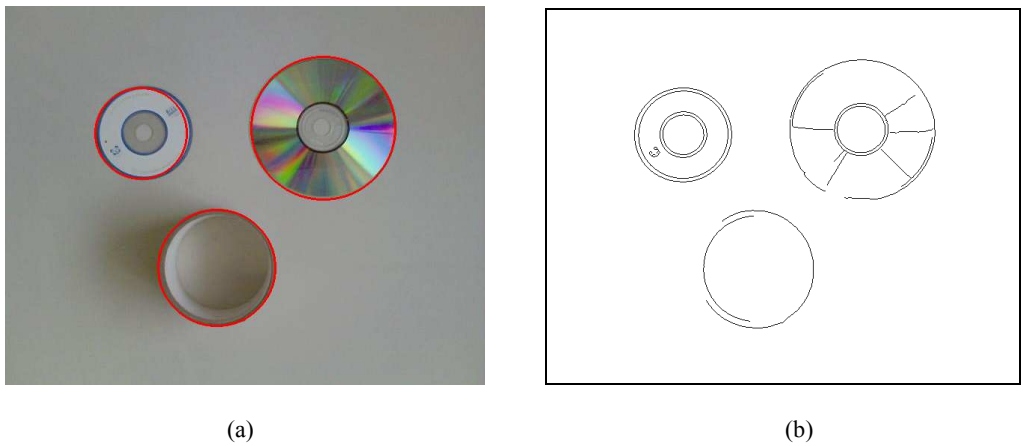

(a)                  (b)

**Fig. 8.** Multiple circle detection on natural images: (a) the original image with an overlay of the detected circles and (b) edge image generated by the canny algorithm prior to be considered by the LA algorithm.

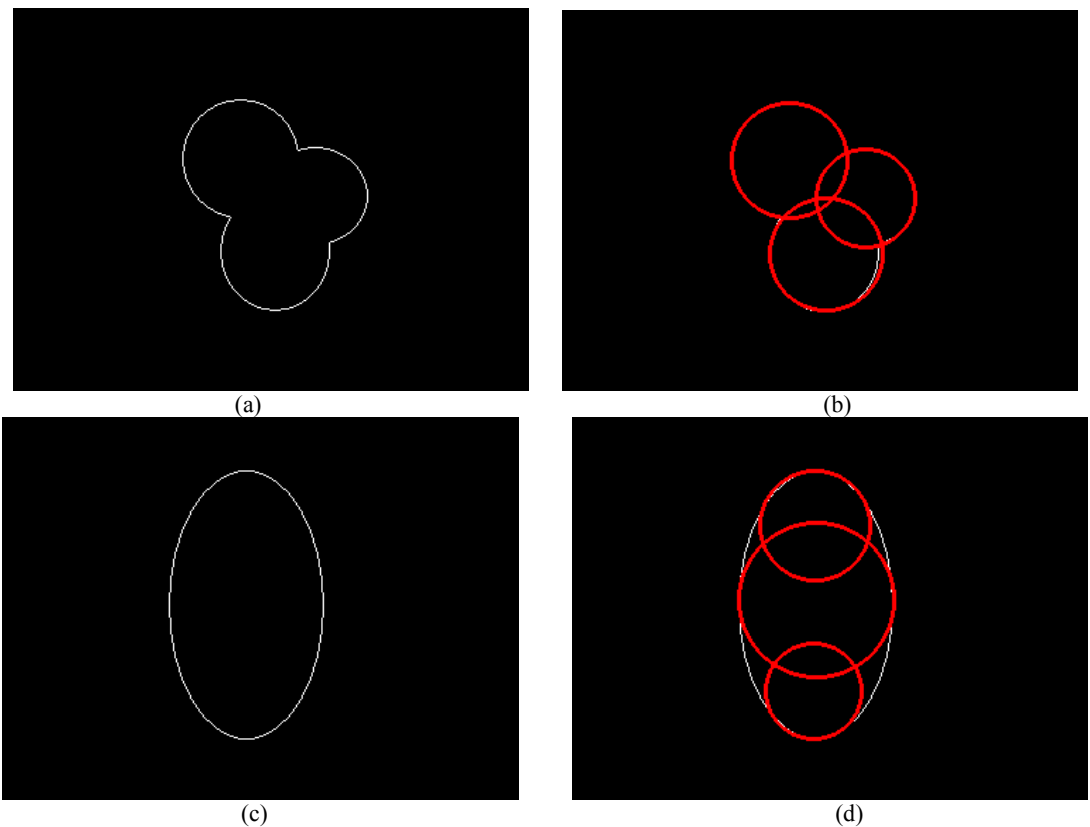

(a)                  (b)

(c)                  (d)

**Fig. 9.** Approximating several shapes by means of circle concatenation: (a)-(c) original images, (b)-(d) their circle approximation.

Figure 9 shows the approximation over several shapes by means of the circle concatenation. In particular Figure 9b shows the circular approximation of a partial circular shape and Figure 9d presents the circular approximation for an ellipse. For both cases, three circles are used to approximate the original shape. Additionally, Figure 10 shows the circular approximation considering a real-life image.

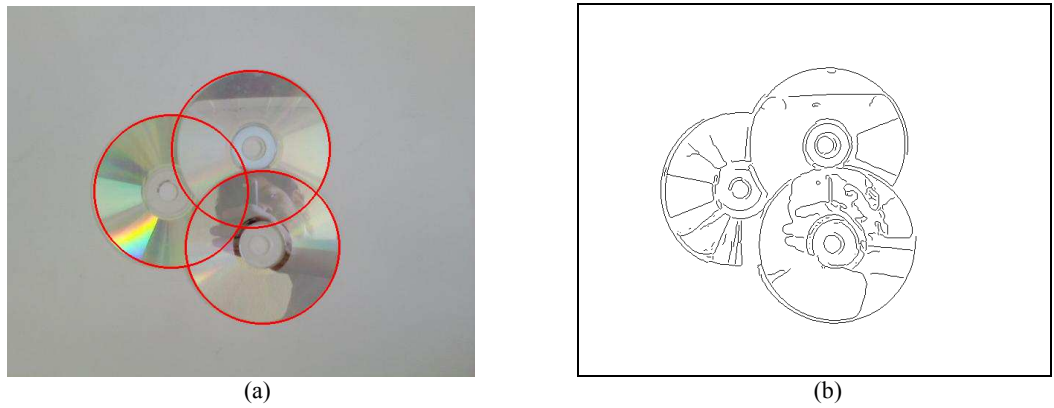

**Fig. 10.** Circular approximation on real-life images: (a) the original image with the detected circles and (b) its edge map.

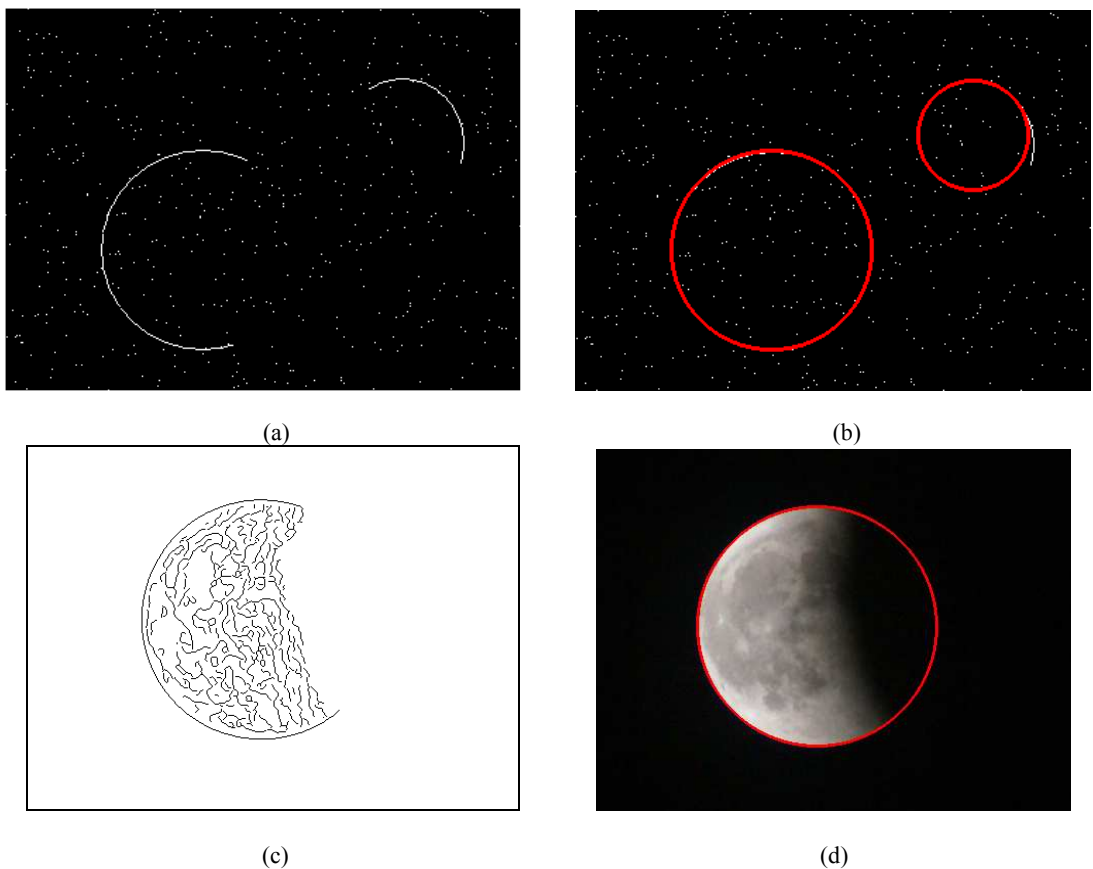

**Fig. 11.** Occluded circles and arc detection: (a) Original synthetic image with two arcs, (b) its circle detection, (c) edge map of a natural occluded circle (the moon), (d) its detected circle as an overlay.

 

*4.5 Occluded circles and arc detection*

The LA circle detector algorithm is also able to detect occluded or imperfect circles as well as partially defined shapes such as arc segments. The LA algorithm achieves the shape matching according to the probability value $p_i(k)$ which represents a score value for a given shape candidate. Figure 11 shows two examples of occluded circles and arcs detection.

*4.6 Complex cases*

In order to test the robustness of the LA algorithm, a particular set of images containing impulsive noise is prepared. They also include other added shapes which can be considered as distracters. Figure 12 presents the results after applying the LA method to such image set.

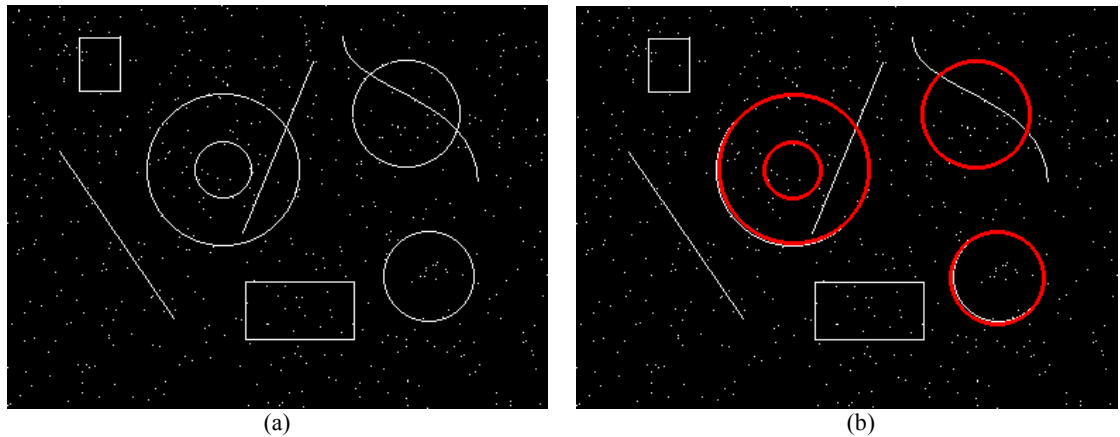

(a)　　　　　　　　　　　　　　　　　　(b)

**Fig. 12.** Circle detection over images with added shapes (distracters) and noise: (a) original image with nine shapes, (b) detection of four circles (overlaid).

Figure 13a shows an image containing six shapes which include three semi-circular patterns. The first circle (top-left on the image) is a quasi-perfect shape while the second (down-right in the image) is an occluded circle. The last circular form has been hand-drawn at the top-right area. Figure 13b shows the image after the circle detection process has been applied.

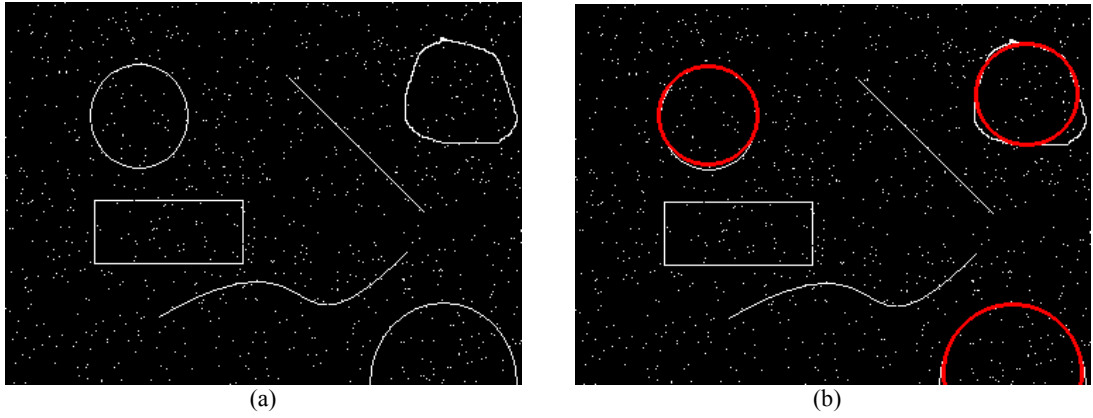

(a)                                                   (b)

**Fig. 13.** Circle detection on images containing different shapes (distracters): (a) Original image including six shapes, (b) Detection of three imperfect circles.

## 5. Performance comparison

In order to enhance the algorithm analysis, the LA algorithm is compared to the GA [12] and the IRHT [7] circle detectors over a set of different images.

*5.1 Parametric setup*

In this comparison, the GA-detector follows the design from Ayala-Ramirez et al. in [12], considering a population size of 70 individuals. The crossover probability, which fixes the threshold for dividing the actual parents' contribution prior to their re-combination into a new individual, is set to 0.55. The mutation probability is considered as 0.10 as it defines the probability of bit-inversion within an individual. Regarding the selection operator, the roulette wheel method is employed. In this approach, each individual is assigned into one slice of the roulette wheel. The size of the slice is proportional to its normalized fitness value. The roulette-wheel strategy favors best fitted individuals but opens a chance of survival for less fitted individuals. The number of elite individuals is set to two, implying that only the best two individuals remain unaltered for the next generation. Moreover, as it is reported in [12], the GA-detector generates the test point set $S = \{s_1(x_1, y_1), s_2(x_2, y_2), \ldots, s_{N_s}(x_{N_s}, y_{N_s})\}$ (used by the fitness function) considering a uniform sampling of the shape boundary by means of the following equations:

$$x_i = x_c + r \cdot \cos\left(\frac{2 \cdot \pi \cdot i}{N_s}\right)$$

$$y_i = y_c + r \cdot \sin\left(\frac{2 \cdot \pi \cdot i}{N_s}\right) \tag{10}$$

where $(x_c, y_c)$, $r$, and $N_s$ represent the circle centre, radius and the desired test-point number, respectively.

The comparative study also includes the IRHT algorithm [7]. For circle/ellipse detection, the IRHT algorithm iteratively applies the randomized Hough transform (RHT) to a region of interest within the image space. Such region is determined from the latest estimation of circle/ellipse parameters $\mathbf{c} = [x_0, y_0, a, b, \phi]$ plus a deviation vector $\mathbf{\Delta_c}$, where $(x_0, y_0)$ are the center coordinates, $a$ and $b$ are the major and minor semi-axes and $\phi$ is the rotation angle. The detection performance of IRHT depends on the values of $\mathbf{\Delta_c}$ which are proportional to the standard error vector $\mathbf{\sigma_c} = [\sigma_x, \sigma_y, \sigma_a, \sigma_b, \sigma_\phi]$ from the latest estimation of $\mathbf{c}$. The comparison considers $\mathbf{\Delta_c} = [0.5 \cdot \sigma_x, 0.5 \cdot \sigma_y, 0.5 \cdot \sigma_a, 0.5 \cdot \sigma_b, 0]$. Such configuration is the least sensitive to noisy images according to [7]. On the other hand, Table 1 summarizes final values for the LA-detector.

*5.2 Error score and success rate*

Images rarely contain perfectly-shaped circles. Therefore, in order to test accuracy, the results are compared to ground-truth circles which are manually detected from the original edge-map. The parameters $(x_{true}, y_{true}, r_{true})$ of the ground-truth circle are computed by using Equations 2-5 considering three circumference points from the manually detected circle. If the centre and the radius of the detected circle are defined as $(x_D, y_D)$ and $r_D$, then an error score can be computed as follows:

$$Es = \eta \cdot (|x_{true} - x_D| + |y_{true} - y_D|) + \mu \cdot |r_{true} - r_D| \tag{11}$$

The first term represents the shift of the centre of the detected circle as it is compared to the benchmark circle. The second term accounts for the difference between their radii. $\eta$ and $\mu$ are two weights which are chosen to agree the required accuracy as $\eta = 0.05$ and $\mu = 0.1$. Such choice ensures that the radii difference would be strongly weighted in comparison to the difference of central circular positions between the manually detected and the machine-detected circles.

In case *Es* is less than 1, the algorithm gets a success; otherwise it has failed on detecting the edge-circle. Notice that for $\eta = 0.05$ and $\mu = 0.1$, it yields *Es*<1 which accounts for a maximal tolerated difference on radius length of 10 pixels, whereas the maximum mismatch for the centre location can be up to 20 pixels. In general, the success rate (SR) can thus be defined as the percentage of reaching success after a certain number of trials.

Figure 14 shows three synthetic images and the results obtained by the GA-based algorithm [12], the IRHT [7] and the proposed approach. Figure 15 presents the same experimental results considering real-life images. The results are averaged over 65 independent runs for each algorithm. Table 2 shows the averaged execution time, the success rate in percentage, and the averaged error score (*Es*) following Equation (10) for all three algorithms over six test images shown by Figures 14 and 15. The best entries are bold-cased in Table 2. A close inspection reveals that the proposed method is able to achieve the highest success rate and the smallest error, still requiring less computational time for most cases.

| Image | Averaged execution time ± Standard deviation (s) | | | Success rate (SR) (%) | | | Averaged *Es* ± Standard deviation | | |
|---|---|---|---|---|---|---|---|---|---|
|  | GA | IRHT | LA | GA | IRHT | LA | GA | IRHT | LA |
| Synthetic images | | | | | | | | | |
| (a) | 2.23±(0.41) | 1.71±(0.51) | **0.21±(0.22)** | 94 | **100** | **100** | 0.41±(0.044) | 0.33±(0.052) | **0.22±(0.033)** |
| (b) | 3.15±(0.39) | 2.80±(0.65) | **0.36±(0.24)** | 81 | 95 | **98** | 0.51±(0.038) | 0.37±(0.032) | **0.26±(0.041)** |
| (c) | 3.02±(0.63) | 4.11±(0.71) | **0.64±(0.19)** | 93 | 78 | **100** | 0.71±(0.036) | 0.77±(0.044) | **0.42±(0.011)** |
| Natural Images | | | | | | | | | |
| (a) | 2.02±(0.32) | 3.11±(0.41) | **0.31±(0.12)** | **100** | **100** | **100** | 0.45±(0.051) | 0.41±(0.029) | **0.25±(0.037)** |
| (b) | 2.11±(0.31) | 3.04±(0.29) | **0.57±(0.13)** | **100** | 92 | **100** | 0.87±(0.071) | 0.71±(0.051) | **0.54±(0.071)** |
| (c) | 2.50±(0.39) | 2.80±(0.17) | **0.51±(0.11)** | 91 | 80 | **97** | 0.67±(0.081) | 0.61±(0.048) | **0.31±(0.015)** |

**Table 2.** The averaged execution-time, the success rate and the averaged *Es* for the GA-based algorithm, the IRHT method and the proposed LA algorithm, considering six test images shown by Figures 14 and 15.

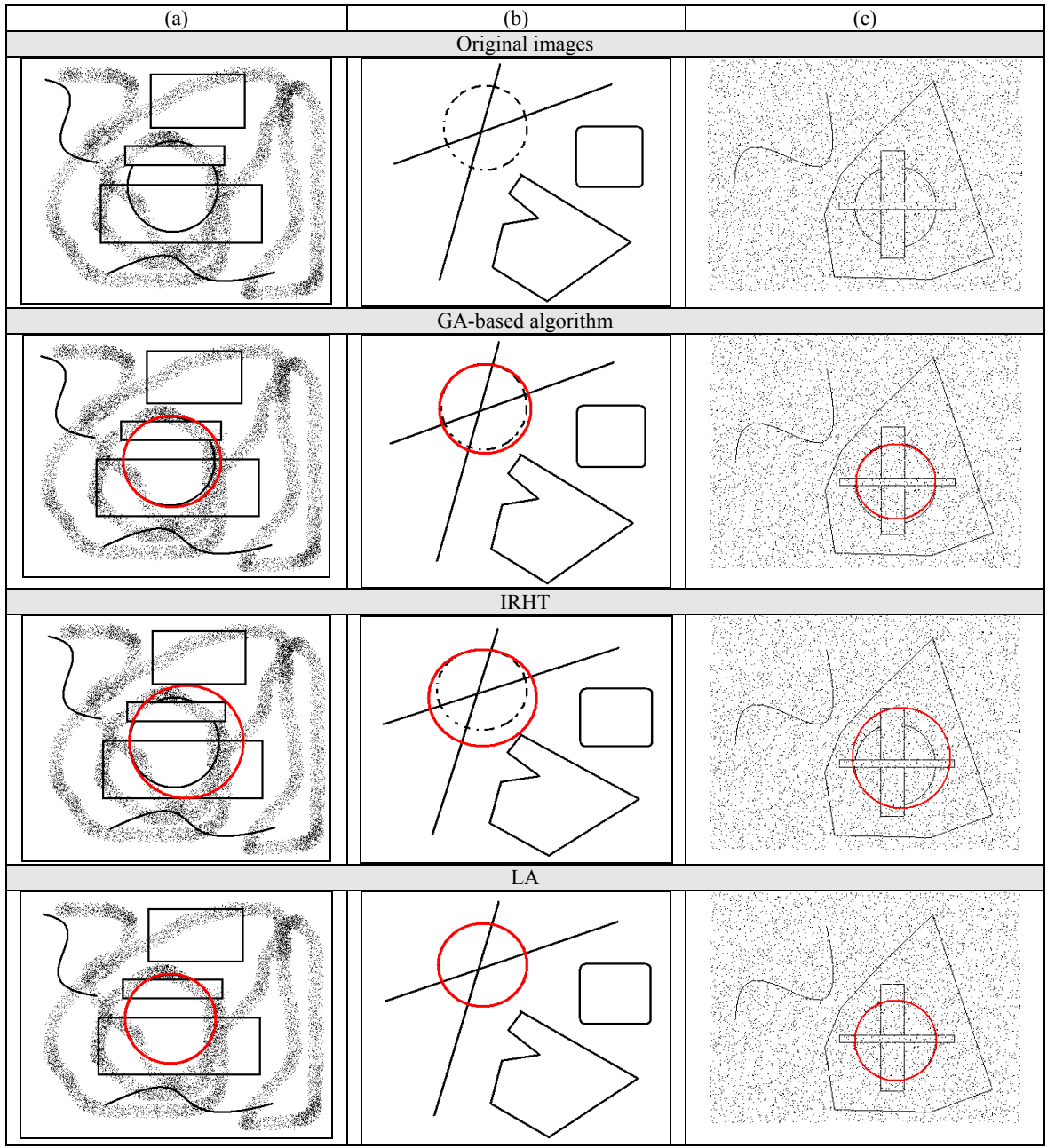

**Fig. 14.** Synthetic images and their detected circles for GA-based algorithm, the IRHT method and the proposed LA algorithm.

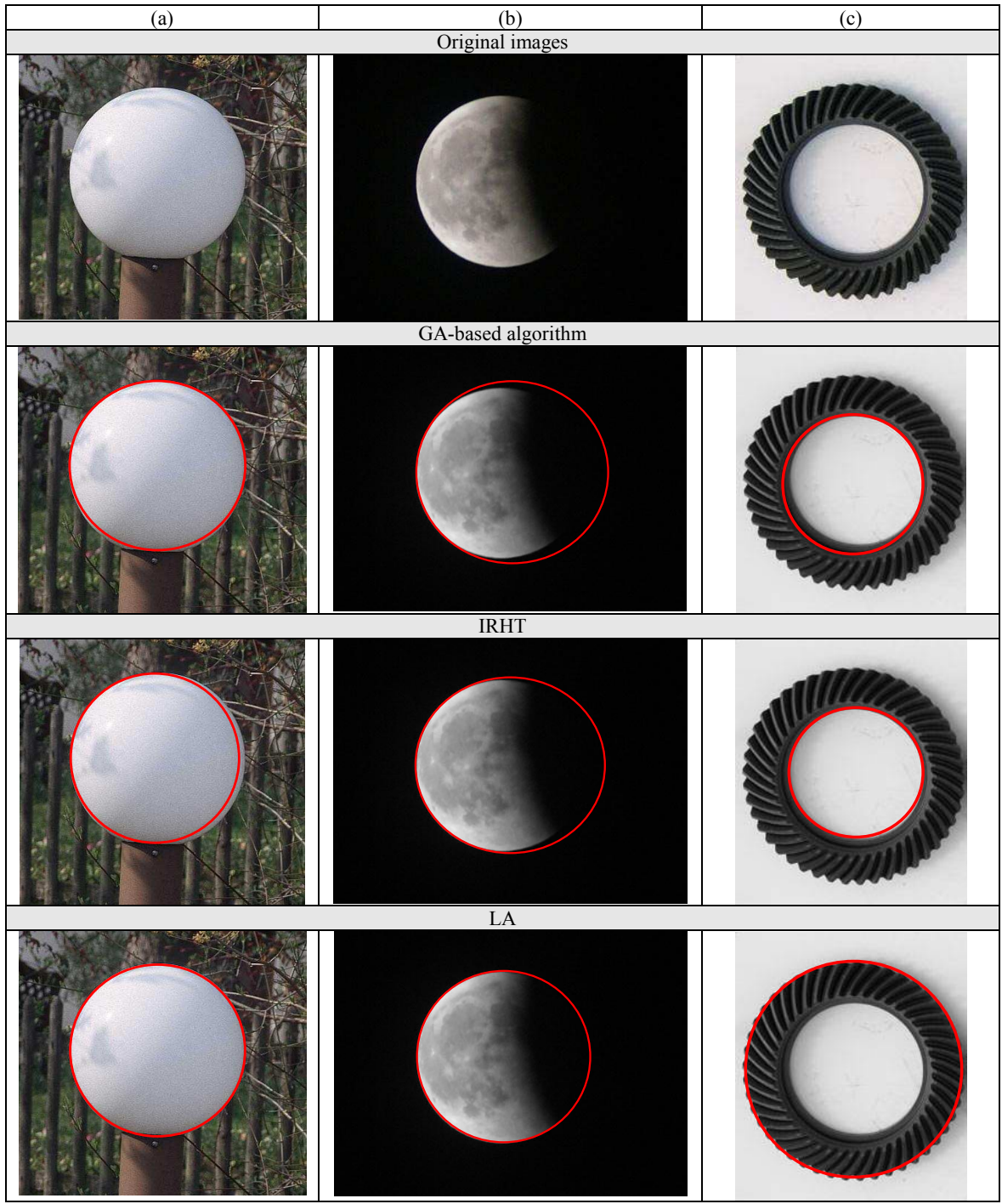

**Fig. 15.** Real-life images and their detected circles for GA-based algorithm, the IRHT method and the proposed LA algorithm.

## 6. Conclusions

This paper has presented an algorithm for the automatic detection of circular shapes from complicated and noisy images with no consideration of the conventional Hough transform principles. The detection process is considered to be similar to an optimization problem. The proposed algorithm is based on Learning Automata (LA) which uses the probability of the three encoded non-collinear edge points as candidate circles (actions) within an edge-only image. A reinforcement signal (matching function) indicates if such candidate circles are actually present in the edge image. Guided by the values of such performance evaluation function, the probability set of the encoded candidate circles are evolved using the LA algorithm so that they can fit into the actual circles (optimal action) in the edge map.

Classical Hough Transform methods for circle detection use three edge points to cast a vote for the potential circular shape in the parameter space. However, they require huge amounts of memory and longer computational times to obtain a sub-pixel resolution. Moreover, the exact parameter set for a detected circle after applying HT frequently does not match the quantized parameter set, rarely finding the exact parameter set for a circle in the image [28]. In our approach, the detected circles are directly obtained from Equations 3 to 6, still reaching a sub-pixel accuracy.

In order to test the circle detection performance, speed and accuracy have been compared. A score function (see Equation (11)) has been proposed to measure the accuracy yielding an effective evaluation of the mismatch between a manually-determined and a machine-detected circle. Moreover, the experimental evidence has demonstrated that the LA method outperforms both the GA (as described in [12]) and the IRHT (as described in [7]) within a statistically significant framework. Table 2 also indicates that the LA method can yield better results on complicated and noisy images in comparison to the GA and the IRHT methods. However, this paper does not aim to beat all the circle detector methods proposed earlier, but to show that the LA algorithm can effectively serve as an attractive method to successfully extract multiple circular shapes.